\documentclass[10pt,twocolumn,letterpaper]{article}

\usepackage{iccv}
\usepackage{times}
\usepackage{epsfig}
\usepackage{graphicx}
\usepackage{amsmath}
\usepackage{amssymb}
\usepackage{color}
\usepackage[dvipsnames]{xcolor}
\usepackage{colortbl}
\usepackage{comment}
\usepackage{placeins}

\usepackage{makecell}
\usepackage{url}
\usepackage{multirow}
\usepackage{floatrow}
\usepackage{listings}
\usepackage{algorithm}
\usepackage{algorithmic}
\usepackage{bm}

\usepackage[accsupp]{axessibility}  %

\usepackage[pagebackref=true,breaklinks=true,letterpaper=true,colorlinks,bookmarks=false]{hyperref}

\definecolor{ggblue}{RGB}{68, 114, 196}
\definecolor{ggred}{RGB}{220, 20, 60}
\definecolor{gggreen}{RGB}{46, 139, 87}
\definecolor{ggorange}{RGB}{237, 125, 49}

\hypersetup{
    colorlinks=true,
    linkcolor=ggred,
    filecolor=ggblue,      
    urlcolor=ggblue,
    citecolor=gggreen,
}

\newcommand\blfootnote[1]{%
  \begingroup
  \renewcommand\thefootnote{}\footnote{#1}%
  \addtocounter{footnote}{-1}%
  \endgroup
}

\newcommand{\shortname}{PARQ\xspace}

\newcommand{\mypar}[1]{\vspace{1mm}\noindent\textbf{#1}}
\newcommand{\myparr}[1]{\vspace{1mm}\noindent\emph{#1}}
\def\eg{\emph{e.g.}\xspace}

\newcommand{\urlNewWindow}[1]{\href[pdfnewwindow=true]{#1}{\nolinkurl{#1}}}

\iccvfinalcopy %

\def\iccvPaperID{5094} %

\ificcvfinal\fi

\begin{document}

\title{Pixel-Aligned Recurrent Queries for Multi-View 3D Object Detection}

\author{
    Yiming Xie$^{1}$ 
    \quad Huaizu Jiang$^{1}$ 
    \quad Georgia Gkioxari$^{*,2}$
    \quad Julian Straub$^{*,3}$
    \\
    $^1$Northeastern University \quad 
    $^2$California Institute of Technology \quad
    $^3$Meta Reality Labs Research\quad 
}

\maketitle
\ificcvfinal\thispagestyle{empty}\fi

\blfootnote{$*$ Equal advising.}

\begin{abstract}
   We present \shortname ~-- a multi-view 3D object detector with transformer and pixel-aligned recurrent queries.
Unlike previous works that use learnable features or only encode 3D point positions as queries in the decoder, \shortname leverages appearance-enhanced queries initialized from reference points in 3D space and updates their 3D location with recurrent cross-attention operations.
Incorporating pixel-aligned features and cross attention enables the model to encode the necessary 3D-to-2D correspondences and capture global contextual information of the input images.
\shortname outperforms prior best methods on the ScanNet and ARKitScenes datasets, learns and detects faster, is more robust to distribution shifts in reference points, can leverage additional input views without retraining, and can adapt inference compute by changing the number of recurrent iterations. 
Code is available at \href{https://ymingxie.github.io/parq}{https://ymingxie.github.io/parq}.

\end{abstract}

\section{Introduction}

The world is composed of objects positioned in 3D space. 
Humans have an innate ability to perceive 3D scenes which allows them to interact with their surroundings. %
For machines, understanding all objects in 3D space from one or few images enables new applications of embodied intelligence such as in robotics and assistive technology. 
The problem is defined by the task of 3D object detection: given a few images of a scene, detect all objects in 3D. 

3D object detection unites two distinct problems of computer vision, 2D recognition, and 3D reconstruction. 
Similar to 2D recognition, appearance cues in the input views drive categorical predictions.
Similar to 3D reconstruction, the model needs to reason about the 3D position of objects from only 2D views. 
Modern learning-based methods build on the traditional multi-view stereopsis~\cite{wheatstone} and Structure from Motion (SfM)~\cite{hartley2003,szeliski2022computer} and lift objects to 3D via optimization~\cite{liFroDODetections3D2020,liMOLTRMultipleObject2020,liODAMObjectDetection,maninisVid2CADCADModel2022} or volumetric representations~\cite{Tulsiani_2017_CVPR,tyszkiewiczRayTran3DPose2022}. 
To do so, they require tens or hundreds of views observing the whole scene. 
However, in many real-world applications, like robotics, models are required to make predictions online and often in real-time from just a short video snippet. 
In this work, we tackle online 3D object detection from just a few (\eg, 3) consecutive views with known camera poses extracted by 
visual-inertial 
SLAM systems~\cite{camposORBSLAM3AccurateOpenSource2020, qinGeneralOptimizationbasedFramework2019a, AugmentedRealityApple}.

\begin{figure}[t]
    \centering
       \includegraphics[width=1.0\linewidth]{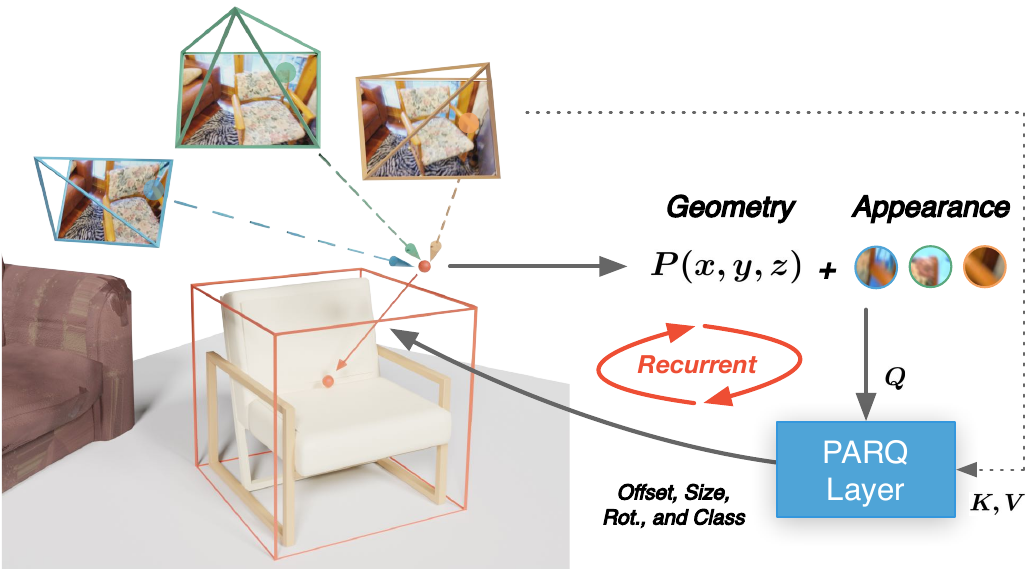}
       \caption{
      \shortname leverages appearance-enhanced queries initialized from 3D points and updates their 3D locations to the 3D object center with a recurrent PARQ layer in the decoder. 
      The PARQ layer chains a transformer decoder layer and a detection head.
        }
       \label{fig:teaser}
\end{figure}

Learning 3D object detectors is challenging as the 3D prediction space is extensive while objects occupy only a small portion.
Geometry is important. 
An object visible in the different input views occupies the same 3D world location and vice versa, a 3D object connects to 2D locations on the input images consistent with its camera projections. 
This observation prunes the vast prediction space.
In addition, object appearance changes with viewpoint changes. 
For instance, the appearance of the chair in Fig.~\ref{fig:teaser} changes as the camera moves. 
This suggests that appearance and geometry are critical.
Motivated by this insight, we design a model that captures geometric and appearance interactions between the 2D input views and the 3D prediction space.

We build on the powerful transformer architecture~\cite{vaswani2017attention}, and specifically DETR~\cite{carionEndtoEndObjectDetection2020}, a popular 2D object detection system. 
DETR encodes input images into feature maps and predicts 2D bounding boxes via cross-attention with learnable queries. 
We enhance DETR in three ways.
First, we make the input feature maps 3D-aware by adding 3D positional information via ray embeddings, following~\cite{liuPETRPositionEmbedding2022}.
Second, we replace the learnable, randomly initialized queries in DETR with appearance- and geometry-informed queries. 
Our queries encode the 3D location of 3D reference points that cover the 3D space. 
They are enhanced with pixel-aligned appearance features sampled from the input views at the projected 2D locations.
Cross-attention between our 3D-aware inputs and appearance-informed 3D queries unleashes our model's ability to capture 3D-to-2D correspondences quickly and efficiently, as we show in our experiments.
Lastly, we deviate from DETR by introducing recurrence. 
DETR makes predictions on the 2D plane while our predictions live in the vast 3D space.
Our initial 3D points are likely to be far from the objects. 
So our model starts by making a coarse prediction that roughly places the initial queries close to true objects and then gradually refines them.
We model this by designing a recurrent scheme that encodes the 3D predictions from the previous step into the current queries.
An overview of our proposed pixel-aligned recurrent queries, dubbed \shortname, is shown in Fig.~\ref{fig:teaser}.

\begin{figure}[t]
    \vspace{-2mm}
    \centering
       \includegraphics[width=1.0\linewidth]{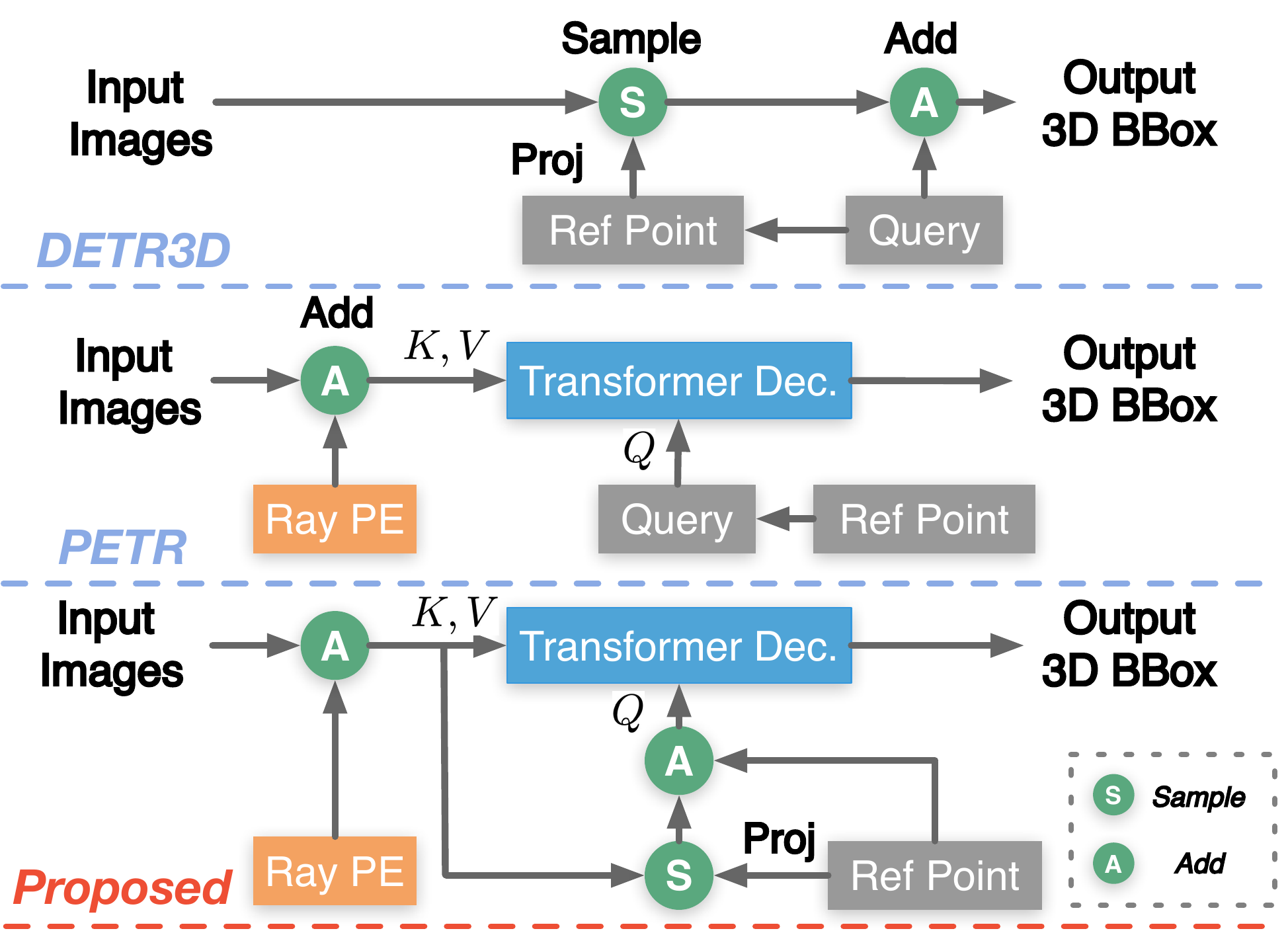}
       \caption{
            \textbf{Query Design.} Schematic comparison of query designs in DETR3D~\cite{wangDETR3D3DObject}, PETR~\cite{liuPETRPositionEmbedding2022}, and our proposed \shortname.
           }
       \label{fig:comparison}
\end{figure}

\shortname differs from prior DETR-style methods for 3D object detection in design and attributes.
Our recurrent decoding and query design differentiates us from the state-of-the-art DETR3D~\cite{wangDETR3D3DObject} and PETR~\cite{liuPETRPositionEmbedding2022}.
DETR3D uses learnable queries and samples local appearance cues from the projected 2D image locations to update the queries. 
This limits the model's ability to capture long-range 3D-to-2D interactions.
PETR enhances the input views with 3D positional information, similar to ours, but only encodes the 3D location of the queries at the start of decoding and without recurrent updates.
This makes it difficult for the model to capture long-range correspondences driven by appearance and forces the model to focus on local cues around the 3D queries, as we show in our experiments.
A schematic comparison is shown in Fig.~\ref{fig:comparison}.
We show that our \shortname outperforms DETR3D and PETR on the challenging ScanNet and ARKitScenes datasets.
More importantly, we show that \shortname exhibits speedier convergence leading to faster training, is robust to a varying number of queries, and can leverage additional input views at test time without the need to retrain. 
In contrast, we show that PETR and DETR3D fail to generalize when deviating from the choice of input views and the number of queries during training.
By tuning the number of queries and recurrent iterations, we show that \shortname detects fastest and most accurately.
Finally, with \shortname we demonstrate zero-shot generalization to novel scenes.

\section{Related Work}

There is a long line of work on 3D object detection from sequences of frames and known camera poses. 
Late fusion techniques detect objects per frame and then fuse these detections, while early fusion methods fuse multi-view features and then detect objects from the fused representation.

\myparr{Late Fusion.} 
In~\cite{liMOLTRMultipleObject2020, liODAMObjectDetection} 3D objects are first detected from single images~\cite{chen2016monocular, brazilKinematic3DObject2020} and then are associated from tens of views via a post-processing optimization step to obtain scene-level predictions.
Dynamic scenes make associations hard as objects move~\cite{cc3dt}. 
Here, velocity estimates can be used to predict associations~\cite{Hu2019Mono3DT}.
In these methods, the quality of the final prediction depends on single-view detection which suffers from scale-depth ambiguity.
To overcome this, another line of work regresses 2D bounding boxes~\cite{heMaskRCNN2018}, and, after association, optimizes 3D quantities to match them. 
\cite{liFroDODetections3D2020} optimizes 3d bounding boxes and shapes to project to associated 2d bounding boxes. 
Similarly, \cite{maninisVid2CADCADModel2022} uses associated 2d detections to instantiate a CAD-model-based reconstruction of a scene via a multi-view constraint optimization formulation.
Common to all late fusion methods is that it is difficult to recover from the errors at the single-view detection and association stage.

\myparr{Early Fusion.}
In early fusion, methods represent geometry either explicitly, \eg, with point clouds, a voxel or BEV grid or, implicitly via input-level inductive biases and positional encodings like ray-encodings~\cite{yifan2022input}.
\cite{rukhovichImVoxelNetImageVoxels2021, tyszkiewiczRayTran3DPose2022, nerfrpn, xu2023nerfdet} adopt a volumetric representation to fuse multi-view image features and predict the 6D pose and the scale of the objects from the feature volume. 
\cite{nerfrpn, xu2023nerfdet} incorporate NeRF~\cite{mildenhall2020nerf} to improve 3D detection but make predictions offline as they assume tens or hundreds of input views.
\cite{yinCenterbased3DObject2020, huangBEVDetHighperformanceMulticamera2022, liBEVFormerLearningBird2022} transform multi-view 2D features into a bird-eye-view (BEV) representation. 
However, volumetric representations require a significant amount of memory which scales with the size of the scene. 
BEV representations, commonly used in urban domains, can mitigate this to some extent by operating in 2D top-down view but cannot represent more complex 3D environments. 
\cite{liuPETRv2UnifiedFramework2022, liBEVFormerLearningBird2022, huangBEVDet4DExploitTemporal2022, wangMonocular3DObject2022} exploit the temporal information across the video to enhance 3D object detection.

\begin{figure*}[ht]
    \centering
       \includegraphics[width=0.95\linewidth]{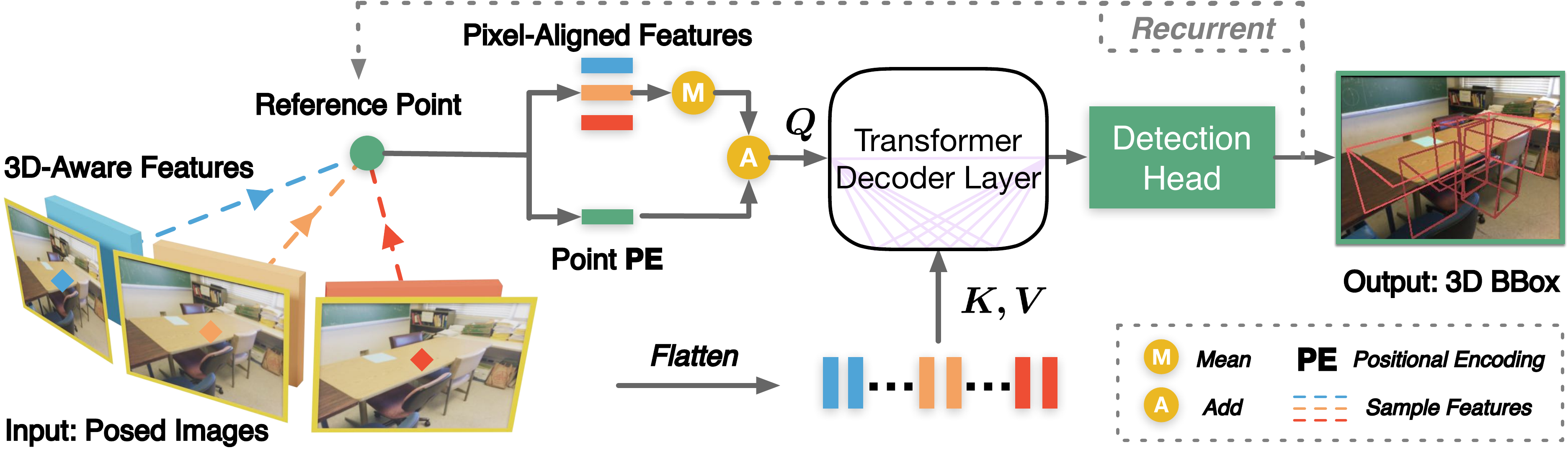}
       \caption{
           \textbf{Overview of \shortname.} Our model predicts 3D object bounding boxes from a short video snippet. We first embed input views with a CNN and add 3D ray-positional encodings. Recurrent \shortname layers consisting of a single Transformer decoder layer and a detection head decode pixel-aligned queries to 3D object predictions. Our queries encode the location of 3D reference points and their appearance cues sampled from the input views.
           Each iteration updates the 3D query points with offsets predicted by the detection head.
           }
       \label{fig:arch}
\end{figure*}

Another line of work explores input-level biases instead of an explicit scene representation.
\cite{wangDETR3D3DObject, liuPETRPositionEmbedding2022, liBEVFormerLearningBird2022} build on DETR~\cite{carionEndtoEndObjectDetection2020} and transformer architectures~\cite{vaswani2017attention,dosovitskiy2020image} where a query represents an object and interacts with the corresponding 2D views to output 3D predictions. 
DETR3D~\cite{wangDETR3D3DObject} predicts 3D objects from learnable randomly-initialized queries which are updated using local appearance cues sampled from the input views.
But when the prediction is far from the true object, relying on local appearance alone is not optimal.
PETR~\cite{liuPETRPositionEmbedding2022} encodes 3D reference points into queries and performs cross-attention with 3D-aware input features.
PETR omits appearance in the queries making convergence slow as the model cannot capture long-range appearance interactions with the 2D input. 
We also adopt a query-based transformer architecture but unlike PETR, we enhance queries with appearance-aligned features.
As a result, our model captures local and global geometric and appearance interactions with the input views.
In addition, through recurrent decoding, the model updates its prediction starting with coarse and then refining its output.

3D object detection is primarily studied in urban and indoor scenes. 
On urban domains, input is captured from camera rigs in autonomous vehicles~\cite{huangBEVDetHighperformanceMulticamera2022, wangDETR3D3DObject, liuPETRPositionEmbedding2022, liBEVFormerLearningBird2022} and methods mainly focus on the \emph{car} category.
In this work, we focus on indoor scenes and inputs from monocular videos.

\section{Method}

In this work, we tackle 3D object detection from a short video snippet, \eg~3 views.
We propose an encoder-decoder architecture that translates geometry- and appearance-informed queries to 3D object detections via recurrent cross-attention with the input views.
Fig.~\ref{fig:arch} shows our architecture.
Given $N$ images from a monocular video %
with known intrinsics and extrinsics, we first extract the image features using a CNN~\cite{heDeepResidualLearning2015}. 
Following~\cite{liuPETRPositionEmbedding2022}, ray position encodings are generated and added to the 2D image features, producing 3D-aware feature maps (Sec.~\ref{2d_features}).
The 3D-aware feature maps interact with point queries via recurrent transformer layers.
Our point queries, anchored on 3D reference points, carry information about their 3D location and appearance as seen from the input views (Sec.~\ref{pixel_aligned_features}).
Via recurrent attention operations with the input views the queries produce the final 3D object detections (Sec.~\ref{decoder}).

\subsection{3D-Aware Input Encoding}
\label{2d_features}
Our model inputs $N$ RGB images $I_i$ with camera parameters $\pi_i$, $i\in \left[1,N\right]$. 
Each image is fed to a ResNet-FPN~\cite{heDeepResidualLearning2015,linFeaturePyramidNetworks2017} which outputs a feature map $F_i \in \mathbb{R}^{H \times W \times C}$. 

We enhance the image features with 3D ray embeddings, following \cite{liuPETRPositionEmbedding2022}.
For each image, we shoot rays originating at the camera center intersecting the image at each pixel. We sample $D$ points along each ray, $P^{ray}_i \in \mathbb{R}^{H \times W \times (D \times 3)}$, with log-scale sampling. More details in the supplementary.

The ray points are transformed to position encodings, $P_i \in \mathbb{R}^{H \times W \times C}$, via an MLP of the same hidden dimension $C$ as the input feature maps.
The ray position encoding, $P_i$, is added to the image feature map, $F_i$, to produce 3D ray-position-aware features, $F_i^{p}=F_i + P_i$. 
The 3D-aware feature maps are input to the transformer decoder. 

\begin{algorithm}[ht]
\small
\caption{\small \textbf{Training Code}
}
\label{alg:train}
\definecolor{codeblue}{rgb}{0.25,0.5,0.5}
\definecolor{codegreen}{rgb}{0,0.6,0}
\definecolor{codekw}{RGB}{207,33,46}
\lstset{
  backgroundcolor=\color{white},
  basicstyle=\fontsize{7.5pt}{7.5pt}\ttfamily\selectfont,
  columns=fullflexible,
  breaklines=true,
  captionpos=b,
  commentstyle=\fontsize{7.5pt}{7.5pt}\color{codegreen},
  keywordstyle=\fontsize{7.5pt}{7.5pt}\color{codekw},
  escapechar={|}, 
}
\begin{lstlisting}[language=python]
def train_loss(images, proj_mat, ref_points, gt, L):
  """
  images: [B, N, H, W, 3]
  proj_mat: projection matrix [B, N, 4, 4]
  ref_points: reference points [K, 3]
  gt: ground truth 3D bounding box [B, ~, 8, 3]
  # B: batch
  # N: number of views
  # K: number of queries
  # L: number of iterations
  """
  
  # Encode image features
  feats = image_encoder(images)

  # Generate ray positional encodings
  ray_encoding = generate_ray_encoding(proj_mat)

  # 3D-aware image features
  feats = feats + ray_encoding

  # supervise the prediction in each iteration
  loss_list = []
  for l in range(L):
    # pixel-aligned features
    pa_feat = sample_and_pool_feats(feats, ref_points, proj_mat)
    
    # point positional encoding
    point_pos = positional_encoding(ref_points)
    
    # transformer decoder layer
    output = layer(tgt=pa_feat, memory=feats, query_pos=point_pos)

    # detection head
    box_param = det_head(output)

    # get the object centers
    box_param['center'] += ref_points
    
    # new reference points for the next iteration
    ref_points = box_param['center'].detach()

    # loss computation
    loss_list.append(loss(box_param, gt))
  
  return loss_list
\end{lstlisting}
\end{algorithm}

\subsection{Appearance- and Geometry-Informed Queries}
\label{pixel_aligned_features}
Now, we turn to our queries.
The purpose of our queries is to produce final 3D object detections by interacting with the input views. 
We randomly generate a set of reference points in the 3D space bounding a large region of the view frustum. 
Assume $x$ is a 3D point and $\pi_i(x)$ is its 2D projection on the $i$-th input view using known camera pose and projection $\pi_i$.
We bilinearly sample the feature vectors at the 2D pixel locations, namely $f_{i,x} = F^p_i(\pi_i(x))$.
We aggregate the feature vectors across all views with average pooling -- if $x$ projects outside the image border for a view it is omitted.
The query for reference point $x$ is defined as:
\begin{equation}
q_x = \textstyle \textrm{MLP}(\gamma(x)) + \frac{1}{N}\sum_i f_{i,x} \,,
\end{equation} 
MLP is a small neural net that embeds the Fourier positional encoding of $x$, $\gamma(x)$~\cite{vaswani2017attention, mildenhallNeRFRepresentingScenes2020}.

By encoding the 3D location of the reference point $x$ and its appearance from the input views, query $q_x$ carries both appearance and geometric cues.
Our appearance- and geometry-informed queries attend to the 3D-aware input feature maps which allows the model to encode 3D-to-2D correspondences and make 3D predictions from just 2D input views. 
This is unlike PETR~\cite{liuPETRPositionEmbedding2022}, where queries encode only the coordinates of the reference points.
We show in our experiments that the absence of appearance cues in the queries leads to lower performance and slower convergence rates.
Note that PIFu~\cite{saitoPIFuPixelAlignedImplicit2019} also remarked the importance of pixel-aligned features when making 3D predictions for the task of human shape reconstruction from images.

\subsection{Recurrent Query-Based Decoding}
\label{decoder}
The goal of the decoder is to predict how a 3D reference point, encoded by the query, should be translated in order to match a true object center.
Initially, the reference point is far from the object, as it is randomly sampled from the 3D space. 
In order to predict how to correctly translate it, we need to capture long-range contextual and geometric cues with the input views. 
We achieve this via cross-attention between our proposed appearance- and geometry-informed queries and 3D-aware input feature maps.

Our decoder is one recurrent \shortname layer which chains a single DETR~\cite{carionEndtoEndObjectDetection2020} transformer decoder layer and a detection head (see Sec.~\ref{heads}). 
The pixel-aligned queries cross-attend to the image features through the multi-head attention operation inside the DETR transformer decoder layer. 
Since the reference point is initially far from the object, we perform $L$ iterative predictions via recurrent decoding.
The reference points are initially randomly positioned in 3D space.
They are recurrently updated with offsets predicted by the \shortname layer. If $x_0$ is the initial 3D reference point,
\begin{align}
    x_l = \text{PARQ}\left(x_{l-1}, \{F_i^p\}\right) + x_{l-1}
\end{align}
Note that all $L$ updates share the same weights in the transformer decoder.
This is unlike DETR~\cite{carionEndtoEndObjectDetection2020} and PETR~\cite{liuPETRPositionEmbedding2022} where distinct non-shared layers in the transformer decoder make object predictions.
In addition to sharing weights across updates, we differ from PETR~\cite{liuPETRPositionEmbedding2022} as we encode the newly predicted object location and its appearance in the query of the subsequent update.
PETR only encodes the initial location of the reference point, $x_0$.
Our experiments show that our design choice is more effective and robust to distribution shifts when sampling reference points.

\subsection{3D Detection Head and Objective}
\label{heads}
The detection head inputs the output of the decoder and makes 3D predictions.
3D objects are represented as 3D bounding boxes with four parameter groups, each predicted by MLPs:
(1) \emph{center offset}: $[\Delta x, \Delta y, \Delta z] \in \mathbb{R}^3$ represents the relative offset of the object center from the reference points, (2) 
\emph{rotation}: $\mathbf{p} \in \mathbb{R}^6$ represents the continuous 6D~\cite{zhouContinuityRotationRepresentations2018} rotation of the object, 
(3) \emph{object size}: $[\hat{w}, \hat{h}, \hat{l}]$ are the box dimensions log-normalized with category-specific pre-computed means,  
and (4) \emph{object class}: $c \in \mathbb{R}^{|\mathcal{C}|+1}$ are confidence scores across the object classes $\mathcal{C}$ (+1 for no-object).
More details in the supplementary.

Following DETR~\cite{carionEndtoEndObjectDetection2020}, we match predictions to ground truths using the Hungarian algorithm~\cite{kuhnHungarianMethodAssignment2010}. 
Aside from Hungarian matching, we also match the GT box and the predictions whose corresponding reference points are in close proximity to this GT box ($< 0.2m$), since for two adjacent reference points which have similar queries, they should both detect nearby objects.
We supervise the object detection output in each iteration.
The training objective, $\mathcal{L}$, is a weighted sum of the losses for center offset $\mathcal{L}_o$, rotation $\mathcal{L}_r$, object size $\mathcal{L}_s$ and classification $\mathcal{L}_c$:
\begin{align}
    \mathcal{L}_d = \alpha_o \mathcal{L}_{o} + \alpha_r \mathcal{L}_{r} + \alpha_s \mathcal{L}_{s} + \alpha_c \mathcal{L}_{c}.
\end{align}
$\mathcal{L}_o$ and $\mathcal{L}_s$ is an $L1$ loss, $\mathcal{L}_r$ is an $L2$ loss, and $\mathcal{L}_c$ is the cross entropy loss. 
We set $\alpha_o$, $\alpha_s$, $\alpha_r$ to 5.0 and $\alpha_c$ to 1.0.

\subsection{Implementation Details}
For the image backbone we use ResNet50~\cite{heDeepResidualLearning2015}, pretrained on ImageNet, integrated with a feature pyramid network (FPN)~\cite{linFeaturePyramidNetworks2017}. 
Unless otherwise stated, we use 3-frame snippets ($N$=3), 8 recurrent updates during decoding ($L$=8), and sample 256 reference points. 
We use the same number of queries for baselines~\cite{liuPETRPositionEmbedding2022, wangDETR3D3DObject}.
The dimension of query and image features is 1024. 
In the transformer decoder layer, we use 4 decoder heads, and the feedforward dimension is 768. The dropout rate is 0.1.
The camera coordinates of the middle snippet frame define the snippet coordinate system.
All 3D predictions are defined with reference to that snippet coordinate system. 
We define the bounding region to sample 3D points of size $6m \times 2.5m \times 5m$ aligned with the snippet coordinate system. 
This volume contains $93.7\%$ of training box centers on ScanNet.
We implement our model using PyTorch~\cite{Paszke_PyTorch_An_Imperative_2019} and train across 8 NVIDIA A5000 GPUs with a batch size $b = 16$ ($2$ per GPU). 
We use the AdamW optimizer~\cite{loshchilov2017decoupled} and an initial learning rate of $10^{-4}$. 
We scale the learning rate by $b/256$~\cite{goyal2017accurate} and use a cosine annealing schedule~\cite{loshchilov2016sgdr}.
Algorithm~\ref{alg:train} provides the pseudo-code for \shortname's training procedure.
See more implementation details in the supplementary.
\section{Experiments}
\begin{table*}[t]
\centering
\resizebox{0.85\textwidth}{!}{
\begin{tabular}{lccccccccc>{\columncolor[gray]{0.902}}c}
\Xhline{3\arrayrulewidth}
@IoU $>$ 0.25 & chair          & table          & cabinet        & trashbin      & bookshelf      & display        & sofa           & bathtub        & other          & \textbf{average}           \\
\hline
ODAM~\cite{liODAMObjectDetection}          & 50.6 & 42.5 & 9.3     & 32   & 19.9 & 14.8  & 39.8 & 28.5 & 0.0          & 33.0/47.1/38.8   \\
ImVoxelNet~\cite{rukhovichImVoxelNetImageVoxels2021}    & 66.0   & \textbf{55.8} & 44.2 & \textbf{52.8} & 13.0    & 0.0          & \textbf{48.1} & 23.3 & \textbf{31.9} & \textbf{55.2}/\textbf{48.6}/\textbf{51.7} \\
DETR3D~\cite{wangDETR3D3DObject}        & 51.6 & 35.4 & 30.4 & 22.4   & 18.5 & 15.6  & 34.8 & 19.1 & 11.2  & 24.4/44.7/31.6 \\
PETR~\cite{liuPETRPositionEmbedding2022}          & 71.0 & 49.3 & \textbf{46.4} & 46.6 & \textbf{29.0} & 26.5 & 44.4 & 40.2 & 23.2 & 49.6/50.5/50.0 \\
\shortname (ours)          & \textbf{72.5} & 46.2 & 44.3 & 51.8 & 20.4 & \textbf{30.6} & 40.5 & \textbf{46.8} & 21.6 & 54.2/48.2/51.1 \\
\Xhline{3\arrayrulewidth}
@IoU $>$ 0.5 & chair          & table          & cabinet        & trashbin      & bookshelf      & display        & sofa           & bathtub        & other          & \textbf{average}           \\
\hline
ODAM~\cite{liODAMObjectDetection}          & 22.5 & 9.9   & 5.0     & 7.6     & 4.8    & 2.7    & 16.2   & 6.7    & 0.0          & 12.1/17.3/14.2 \\
ImVoxelNet~\cite{rukhovichImVoxelNetImageVoxels2021}    & 43.8 & 17.5   & 18.8 & \textbf{21.5}   & 0.8    & 0.0          & 16.7   & 12.3  & \textbf{13.8} & 28.6/25.2/26.8       \\
DETR3D~\cite{wangDETR3D3DObject}        & 25.4 & 6.7      & 11.6 & 3.7     & 3.0      & 2.1    & 12.0    & 4.5    & 1.9    & 8.8/16.1/11.4  \\
PETR~\cite{liuPETRPositionEmbedding2022}          & 44.2 & 20.5 & 25.8 & 12.7 & \textbf{9.3}   & 5.1    & \textbf{22.2} & 16.8 & 9.3   & 25.4/25.9/25.6 \\
\shortname (ours)          & \textbf{52.0} & \textbf{20.7} & \textbf{27.9} & 18.3 & 6.0   & \textbf{7.1}   & 19.0 & \textbf{20.9} & 9.9   & \textbf{31.8}/\textbf{28.3}/\textbf{30.0} \\
\Xhline{3\arrayrulewidth}
@IoU $>$ 0.7 & chair          & table          & cabinet        & trashbin      & bookshelf      & display        & sofa           & bathtub        & other          & \textbf{average}           \\
\hline
ODAM~\cite{liODAMObjectDetection}          & 2.3    & 0.6    & 0.0          & 1.6    & 0.0          & 0.0          & 1.0      & 0.8    & 0.0          & 1.2/1.7/1.4    \\
ImVoxelNet~\cite{rukhovichImVoxelNetImageVoxels2021}    & 6.7    & 1.8    & 2.2    & 0.9    & 0.0          & 0.0          & 3.7    & \textbf{2.7}    & \textbf{1.7}    & 4.1/3.6/3.8      \\
DETR3D~\cite{wangDETR3D3DObject}        & 3.2    & 0.2    & 0.4    & 0.0          & 0.0          & 0.2    & 0.0          & 1.3      & 0.0          & 0.9/1.7/1.2    \\
PETR~\cite{liuPETRPositionEmbedding2022}          & 9.1   & 2.0    & \textbf{7.7}    & 0.4    & 0.1    & 0.0          & 1.2    & 1.9    & 1.4    & 4.6/4.6/4.6    \\
\shortname (ours)          & \textbf{11.8} & \textbf{2.9}    & 7.0    & \textbf{2.0}    & \textbf{1.2}    & \textbf{0.4}    & \textbf{3.8}    & 2.0    & 1.6    & \textbf{6.5}/\textbf{5.8}/\textbf{6.1}   \\
\Xhline{3\arrayrulewidth}
\end{tabular}
}
\caption{\textbf{Performance on ScanNet}. 
We compare \shortname to prior works ODAM~\cite{liODAMObjectDetection}, ImVoxelNet~\cite{rukhovichImVoxelNetImageVoxels2021}, DETR3D~\cite{wangDETR3D3DObject} and PETR~\cite{liuPETRPositionEmbedding2022}.
We report F1 for each of the 9 classes of ScanNet and Precision/Recall/F1 for average performance.
Note that ODAM omits the `other' class, so we exclude it from the average.
See full Precision/Recall/F1 performance for all classes in the supplementary.
    }
\label{tab:scannet_view3}
\end{table*}
\begin{table}[t]
    \centering
    \resizebox{1.0\columnwidth}{!}{
        \begin{tabular}{lccccc}
            \Xhline{3\arrayrulewidth}
            Prec./Rec./F1                                              & @IoU$>$0.25                    & @IoU$>$0.5      & @IoU$>$0.7  \\
            \hline
            ImVoxelNet~\cite{rukhovichImVoxelNetImageVoxels2021}  & 44.5/40.3/42.3      & 15.9/14.4/15.1        &      1.3/1.2/1.3                    \\
            PETR~\cite{liuPETRPositionEmbedding2022}  &  36.6/\textbf{53.2}/43.4                               & 14.8/21.5/17.5                &  2.5/3.7/3.0      \\
            Ours  &  \textbf{54.1}/44.4/\textbf{48.8}             & \textbf{26.7}/\textbf{21.9}/\textbf{24.1}   &  \textbf{6.3}/\textbf{5.2}/\textbf{5.7}    \\ 
            \Xhline{3\arrayrulewidth}
        \end{tabular}}
        \caption{
        \textbf{Performance on ARKitScenes} with 3 views. We report average Precision/Recall/F1 for all 17 classes. 
        The complete table is provided in the supplementary.
        }
        \label{tab:arkit}
\end{table}

We tackle 3D object detection from a few (\eg~3) consecutive RGB frames of complex indoor scenes with many object types. 
We provide an extensive quantitative and qualitative analysis of our model's performance and show that our approach outperforms the previous state-of-the-art methods for the task. 
More importantly, we show that appearance-informed queries lead to better performance, faster convergence and can leverage more input views during inference.
Finally, we test our model's generalization ability by deploying it on user-captured videos without any finetuning. There, scenes are captured with an iPhoneXR, and camera poses are obtained from ARKit~\cite{AugmentedRealityApple}. 

\mypar{Datasets.}
We experiment on the popular ScanNetv2~\cite{daiScanNetRichlyannotated3D2017} and ARKitScenes~\cite{arkitscenes} datasets. The ScanNet dataset contains RGB-D videos of 1613 indoor scenes with multiple objects in complex spatial arrangements.
Scan2CAD~\cite{avetisyanScan2CADLearningCAD2018} aligns CAD models which are used to extract 3D bounding box annotations for all objects in the scene.
Following~\cite{tyszkiewiczRayTran3DPose2022}, we evaluate on 9 classes
and use the official train/val splits by~\cite{daiScanNetRichlyannotated3D2017}.
On ScanNet, the input image size is $320 \times 240$ and the image feature map size is $80 \times 60$. 
ARKitScenes is a challenging dataset of indoor scenes which includes manually labeled 3D oriented bounding boxes for a large taxonomy of furniture. 
We follow the official train/val/test split and evaluate for all 17 classes.
The orientation of the images in the ARKitScenes dataset vary from video to video. 
Even though they sky direction of each video is provided in the metadata, some labels are inaccurate \footnote{\href{https://github.com/apple/ARKitScenes/issues/10}{https://github.com/apple/ARKitScenes/issues/10}}. 
We follow the metadata to rectify the images. %
We use the videos with sky directions `Up' and `Down'.
On ARKitScenes, the input image size is $256 \times 192$ and the image feature map size is $64 \times 48$.

\begin{figure*}[ht]
    \vspace{-0.5cm}
    \centering
       \includegraphics[width=1\linewidth]{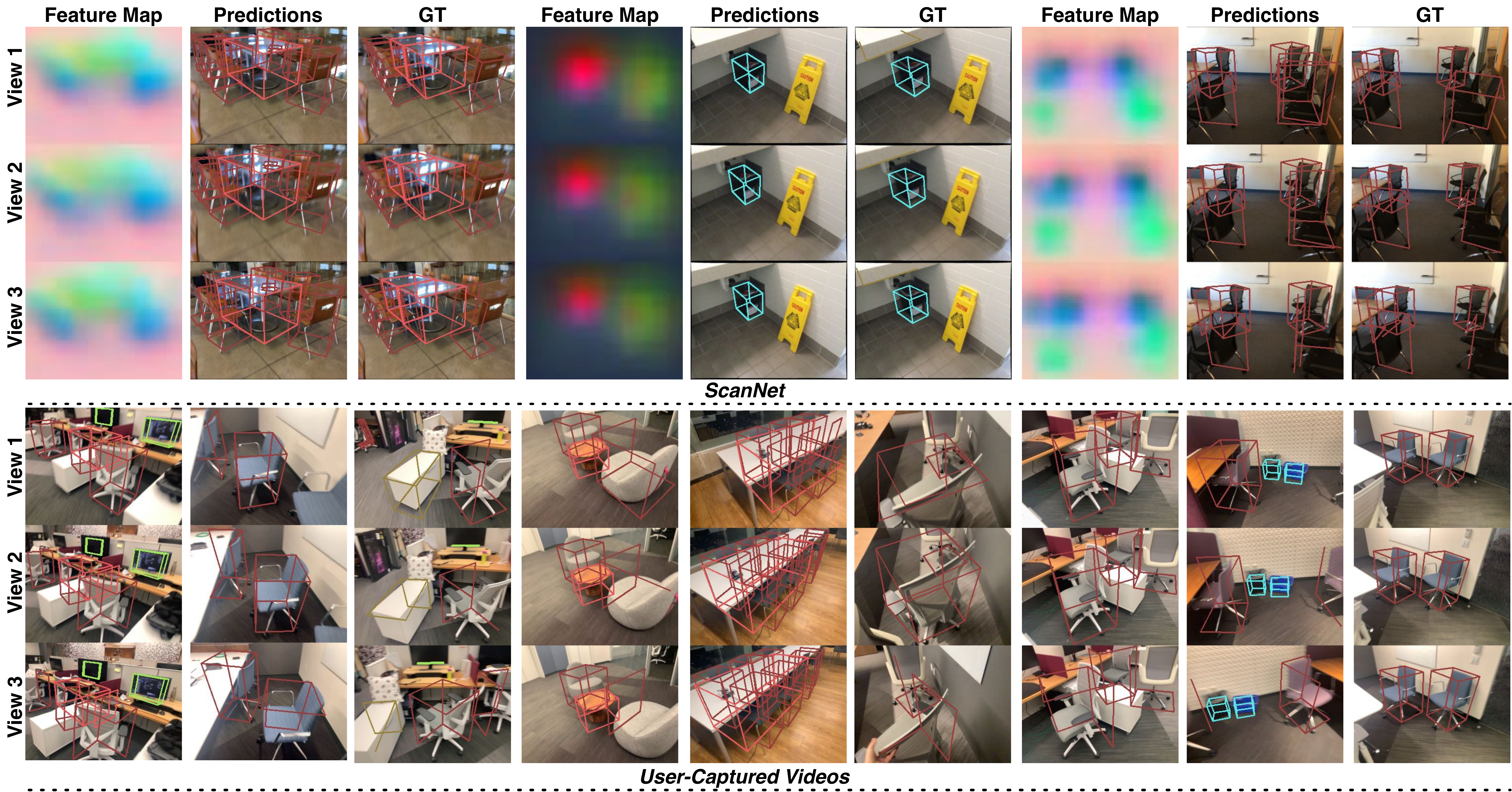}
       \caption{
           \textbf{Qualitative results on ScanNet (top) and user-captured videos (bottom).}
        \emph{Zoom in for details}. 
        We compress the image feature maps using Linear PCA~\cite{halko2011finding}.
        Note that the learned feature maps are multi-view consistent. 
        We also test our model’s generalization ability by deploying it on user-captured videos without any fine-tuning. 
        More results in the supplementary.
        }
       \label{fig:qualitative}
    \vspace{-0.2cm}
\end{figure*}

\paragraph{Extracting video snippets.}
We focus on online 3D object detection from a short video snippet. 
Given a monocular video and per-frame camera poses, extracted from~\cite{dai2017bundlefusion,AugmentedRealityApple}, we split the video into snippets of $N$ frames as follows:
The first frame is selected. 
Similar to~\cite{sunNeuralReconRealTimeCoherent2021, xiePlanarReconRealtime3D2022}, the next frame is added if its relative translation is greater than $0.1m$ or its relative rotation angle is greater than $15^o$ compared to the last selected frame. 
Once $N$ frames are selected, they form the snippet.
Our selection process ensures that snippets contain consecutive video frames which are relatively visually diverse.
Examples of our snippets are shown in Fig.~\ref{fig:qualitative}.

Next, we extract 3D box annotations for each snippet. 
Both Scan2CAD and ARKitScenes provide 3D annotations for the entire scene in world space, which is not suitable for us.
For each snippet, we keep annotations belonging to the \emph{visible} objects in the snippet.
To determine whether an object is within the snippet frustum, we project the corners of its 3D box on the views and calculate the IoU between the projected 2D box and the image border. If the IoU is less than $0.5$ we remove the box.
To determine whether an object is severely occluded in the snippet views, we un-project the ground-truth view depth maps and calculate the number of points inside the 3D box. If the number of points is less than $100$ we remove the box.

\paragraph{Metrics.}
For all experiments, we adopt the popular evaluation protocol from ODAM~\cite{liODAMObjectDetection} which evaluates at the scene level.
For each scene, we keep the predictions above a confidence threshold $s$ from each input snippet.
Hungarian matching~\cite{kuhnHungarianMethodAssignment2010} with 3D Intersection-over-Union (IoU) is used to match predictions between the current and the previous snippet.
For two matched boxes, we only keep the box with the higher score as one of the scene-level predictions.
3D NMS is used to filter out potential duplicate predictions.
More details in the supplementary.
We use the same tracking and fusion strategy for baselines and our model for a fair comparison. 

A prediction is considered a true positive if its 3D IoU with a ground-truth of the same class is above a predefined threshold $\tau$.
Duplicate predictions, namely predictions paired to an already matched true object, are marked as false positives. 
We report Precision, Recall, and F1:
{\small
\begin{align}
    & Prec. = \frac{N_{tp}}{N_{pred}},~~ Rec. = \frac{N_{tp}}{N_{gt}}, ~~F1 = \frac{2 * Prec. * Rec.}{Prec. + Rec.}
\end{align}
}%
$N_{tp}$, $N_{pred}$, $N_{gt}$ are the numbers of true positives, predictions, and true objects.
We set $\tau = [0.25, 0.5, 0.7]$. 

\subsection{Comparison to Other Methods}

We compare our method to recent multi-view 3D object detection models.
Two adopt a DETR-style approach, one is a volumetric approach and one is a late fusion method.
\textbf{DETR3D}~\cite{wangDETR3D3DObject} follows DETR~\cite{carionEndtoEndObjectDetection2020} and uses learnable queries to predict the 3D bounding boxes. The learnable queries in DETR3D predict the reference points which are then used to sample local appearance cues from the input views at the projected 2D locations and update the queries.
\textbf{PETR}~\cite{liuPETRPositionEmbedding2022} adopts a DETR architecture where queries encode the 3D location of the initial reference points and are then used to attend to 3D-aware input feature maps. The queries are updated via transformer layers. 
Our approach differs from PETR in two distinct ways: (1) our queries encode both the 3D location and local appearance features of the 3D reference points, (2) we adopt recurrent decoding with attention to refine the 3D reference points. The refined location along with its corresponding appearance features is encoded by the queries for the next recurrent layer.
\textbf{ImVoxelNet}~\cite{rukhovichImVoxelNetImageVoxels2021} fuses the input views into a voxel representation which is then used to cast 3D predictions.
Note that ImVoxelNet predicts axis-aligned 3D bounding boxes on ScanNet.
We retrain DETR3D, PETR, and ImVoxelNet to take as input the same views as our model and predict oriented 3D boxes.
We use the same 2D image backbone and detection head for DETR3D and PETR and follow the official implementation of Transformer architecture.
The voxel grid in ImVoxelNet is shaped $64 \times 64 \times 32$ with a $0.08m$ voxel unit.
Finally, we compare to \textbf{ODAM}~\cite{liODAMObjectDetection} a late fusion approach which processes single-view detections from all frames of a snippet by optimizing a multi-view objective. 
All methods are evaluated identically following our evaluation protocol described above.
We select the confidence threshold $s$ for each method so as to maximize each model's F1 score.

Table~\ref{tab:scannet_view3} compares performance on ScanNet with 3-frame snippets.
We report the F1 score for each object class and Prec./Rec./F1 for the average performance. 
The complete table is provided in the supplementary.
We highlight some interesting observations.
Our method outperforms all baselines at the stricter IoU thresholds of 0.5 and 0.7, by $+3.2\%$ and $+1.5\%$ F1, respectively.
Compared to the DETR-style baselines~\cite{wangDETR3D3DObject, liuPETRPositionEmbedding2022}, we note that our approach performs best on average for all IoU thresholds proving that capturing long-range appearance and geometry cues via our geometry- and appearance-informed queries is effective. 
Our method surpasses the volumetric method ImVoxelNet~\cite{rukhovichImVoxelNetImageVoxels2021} at the stricter IoU thresholds but ImVoxelNet has a small advantage of $+0.6\%$ F1 at the 0.25 IoU threshold.
Since ImVoxelNet constructs a volume to fuse multi-view features, performance is tied to the voxel resolution. This leads to a competitive performance at the loose IoU threshold (0.25) but worse results at stricter IoU thresholds. 
Increasing the voxel resolution could increase performance, but would result in a cubic increase in memory for the model.
Table \ref{tab:arkit} reports the performance on the ARKitScenes dataset. 
We report Prec./Rec./F1 for the average performance.
Our approach outperforms all baselines across all IoU thresholds in F1. 
We provide the complete table in the supplementary.

\subsection{Ablation Study}
We conduct several ablation experiments on ScanNet to validate the effectiveness of our model's design choices, shown in Table~\ref{tab:scannet_view3_ablation}.
We summarize key findings below.

\begin{figure}[t]
    \centering
       \includegraphics[width=1.0\linewidth]{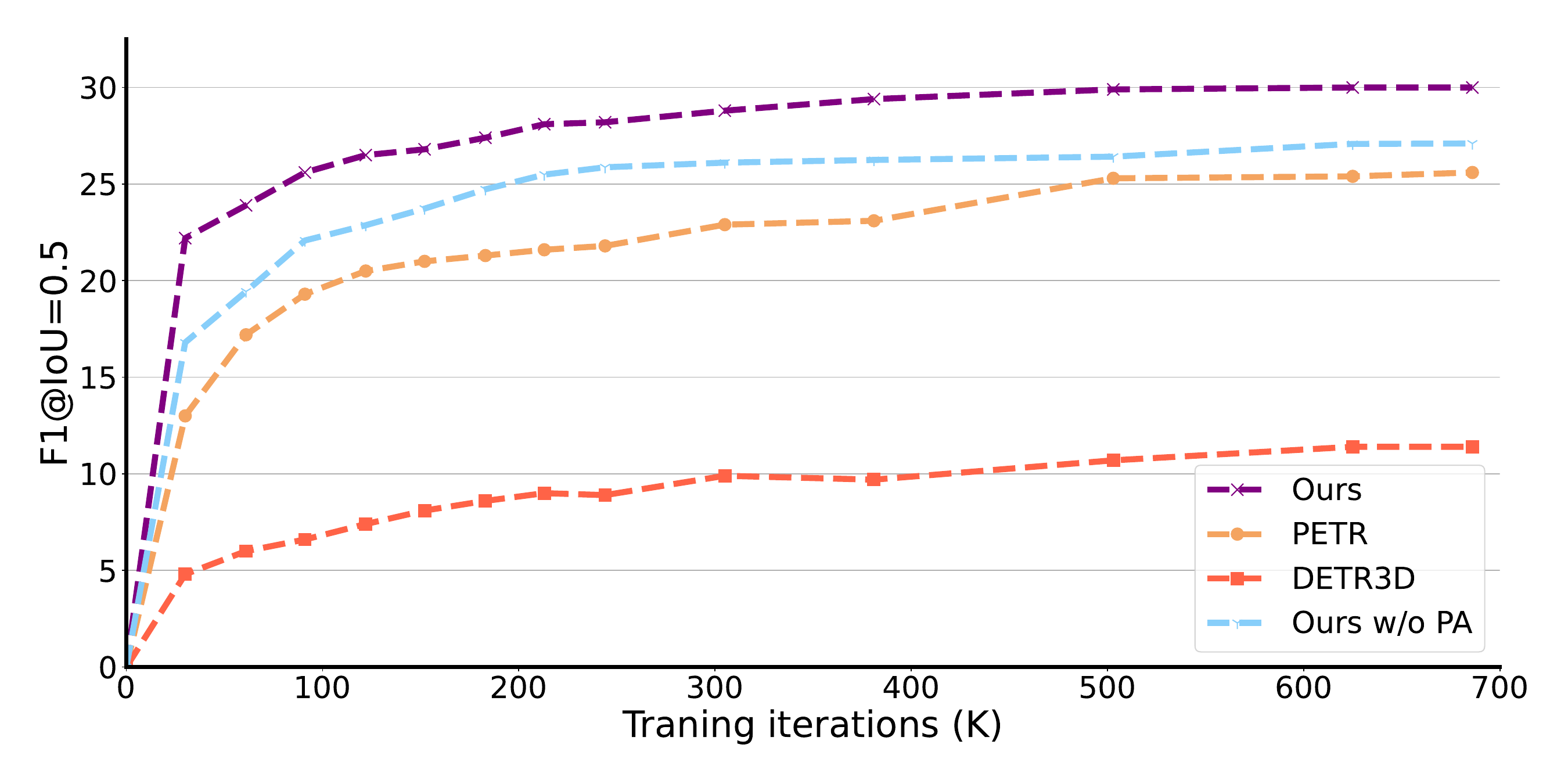}
       \caption{
            \textbf{Convergence speed} for \shortname and competing methods.
           }
       \label{fig:convergency}
    \vspace{-0.1cm}
\end{figure}

\begin{figure*}[ht]
    \centering
       \includegraphics[width=1\linewidth]{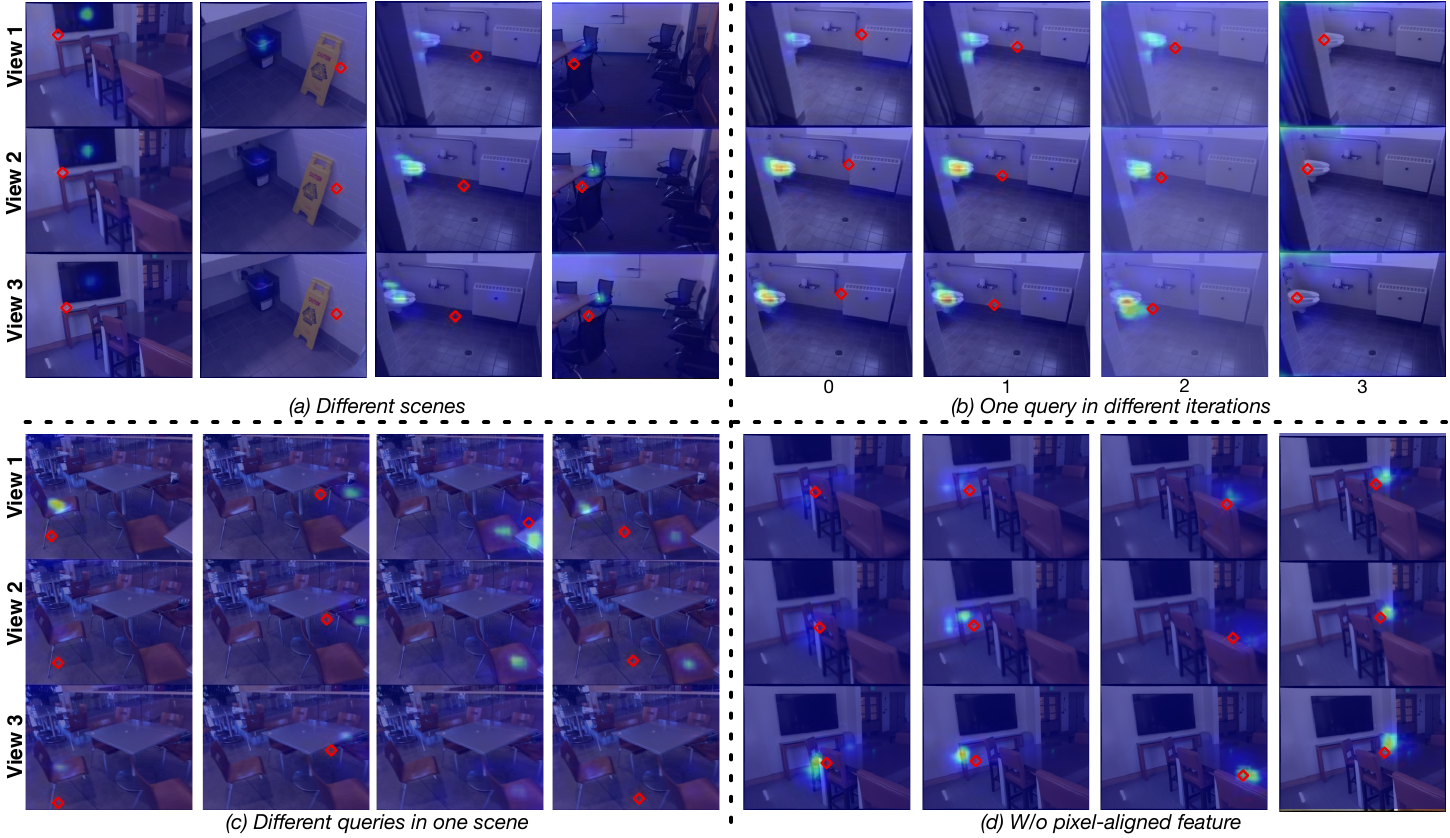}
       \caption{
           \textbf{Attention maps.} \emph{Zoom in for details}.
        We visualize the attention map between queries and multi-view images.
        The \textit{\textcolor{red}{red}} marker represents the projected 3D point.
        (a) Attention for different scenes after \shortname layer 0.
        (b) During the recurrent updates, the 3D point approaches the region with high attention weights.
        (c) Queries for a scene tend to attend to objects closest to them. 
        (d) Without pixel-aligned features, as in PETR, queries focus only on the local area around them; in stark contrast to \shortname in (a-c) which attends globally.
        }
       \label{fig:attention}
    \vspace{-0.2cm}
\end{figure*}

\mypar{Pixel-aligned queries perform better and train faster.}
To show the impact of appearance-informed queries we compare our model (1$^\textrm{st}$ row) to a variant without appearance cues in the queries, \textit{w/o PA} (2$^\textrm{nd}$ row). 
Appearance-informed queries perform better across all IoU thresholds and show a $+2.9\%$ boost in F1 at 0.5 IoU. 
Fig.~\ref{fig:attention} validates this observation by visualizing the attention maps.
Fig.~\ref{fig:attention}(d) qualitatively shows how queries \emph{without} appearance information focus only on the local area around the reference point while our appearance-informed queries can capture long-range contextual interactions in the input views as shown in Fig.~\ref{fig:attention}(a-c).
Additionally, appearance-informed queries lead to faster convergence as shown in Fig.~\ref{fig:convergency}.
These results show that appearance cues in queries are effective.

\definecolor{gray}{rgb}{0.902, 0.902, 0.902}
\newcommand{\CC}{\cellcolor{gray}}

\begin{table}[t]
    \centering
    \resizebox{1.0\columnwidth}{!}{
        \begin{tabular}{lccccc}
            \Xhline{3\arrayrulewidth}
            Prec./Rec./F1                                              & @IoU$>$0.25                    & @IoU$>$0.5      & @IoU$>$0.7  \\
            \hline
            \shortname                                                & 54.2/\textbf{48.2}/\textbf{51.1}                      &  \textbf{31.8}/\textbf{28.3}/\textbf{30.0}      & \textbf{6.5}/\textbf{5.8}/\textbf{6.1}              \\
            \rowcolor[gray]{0.902} w/o PA  &  52.3/48.2/50.2                               & 28.6/25.8/27.1                &  5.4/4.9/5.1      \\
            \rowcolor[gray]{0.902} w/o recurrence  & \textbf{54.3}/47.9/50.9                               & 27.9/27.2/27.6                &  6.0/5.1/5.5                 \\ 
            \rowcolor[gray]{0.902} w/o point PE  &  12.0/22.4/15.6                               &  9.1/12.2/10.4                & 1.0/1.2/1.1              \\
            \rowcolor[gray]{0.902} w/o ray PE  & 53.4/48.0/50.6                               &  29.4/27.1/28.2                & 5.9/5.3/5.6     \\
            \Xhline{3\arrayrulewidth}
        \end{tabular} }
        \caption{
    \textbf{\shortname ablations.} We compare \shortname on ScanNet with variants that remove pixel-aligned features from the  queries (w/o PA), perform no recurrent 3D refinement but updates query features as in PETR (w/o recurrence), remove the positional encoding of 3D points in queries (w/o point PE) and remove the ray positional encoding in image features (w/o ray PE). 
    }
    \label{tab:scannet_view3_ablation}
\end{table}

\mypar{Recurrent refinement improves performance.}
Compared to a variant without recurrent refinement similar to PETR, \textit{w/o recurrence} (3$^\textrm{rd}$ row), our proposed recurrent update strategy gives up to $+2.0\%$ boost in F1. 
Fig.~\ref{fig:attention}(b) visualizes the attention maps after each iteration.
During our updates, the appearance-informed query attends to the target object, even at iteration 0 where the target is furthest away, and moves closer to the target with every update.

Note that PETR updates queries after every transformer layer in the decoder. 
However, the new queries are the output of the previous attention layer and don't encode the newly predicted 3D locations.
This is a stark difference from our approach which updates queries with new appearance and geometry information after each iteration.

\mypar{Geometry-informed queries are critical.}
Embedding the 3D reference points in the queries is critical, as shown by our ablation \textit{w/o point PE} (4$^\textrm{th}$ row). 
This is not a surprise as our model is tasked to predict the target offset from the 3D reference point, so knowledge of its 3D location is helpful.

\mypar{3D-aware input features are helpful.}
We remove ray positional encodings in the input feature maps, proposed by~\cite{liuPETRPositionEmbedding2022}, in \textit{w/o ray PE} (5$^\textrm{th}$ row). 
We notice that ray positional encodings slightly boost performance.

\begin{table}[t]
    \centering
    \resizebox{1.0\columnwidth}{!}{
        \begin{tabular}{lcccccccc}
            \Xhline{3\arrayrulewidth}
            F1         &            1 view &  2 views                         & 3 views                    & 5 views      & 7 views & 9 views  \\
            \hline
            ImVoxelNet~\cite{rukhovichImVoxelNetImageVoxels2021}  & 25.0 & 26.6 &  26.8                 & 26.5                &  25.0    & 23.3  \\
            PETR~\cite{liuPETRPositionEmbedding2022} & 21.3 & 21.7 &  25.6                               & 24.5                &  24.4    & 21.0  \\
            \shortname (ours)                         & \textbf{26.3} &       \textbf{27.6}                   & \textbf{30.0}                      &  \textbf{31.1}      & \textbf{31.3}  & \textbf{31.1}           \\
            \Xhline{3\arrayrulewidth}
        \end{tabular} }
        \caption{\textbf{Number of views}. We test models on different number of views. We train with 3 views and report F1 at @IoU $>$ 0.5.}
    \label{tab:views}
\end{table}

\subsection{Deeper Dive into \shortname}
Next, we investigate the properties of \shortname to better understand its characteristics and highlight its strengths. 
The experiments are conducted on the ScanNet dataset. 

\begin{figure}[t]
    \centering
       \includegraphics[width=1.0\linewidth]{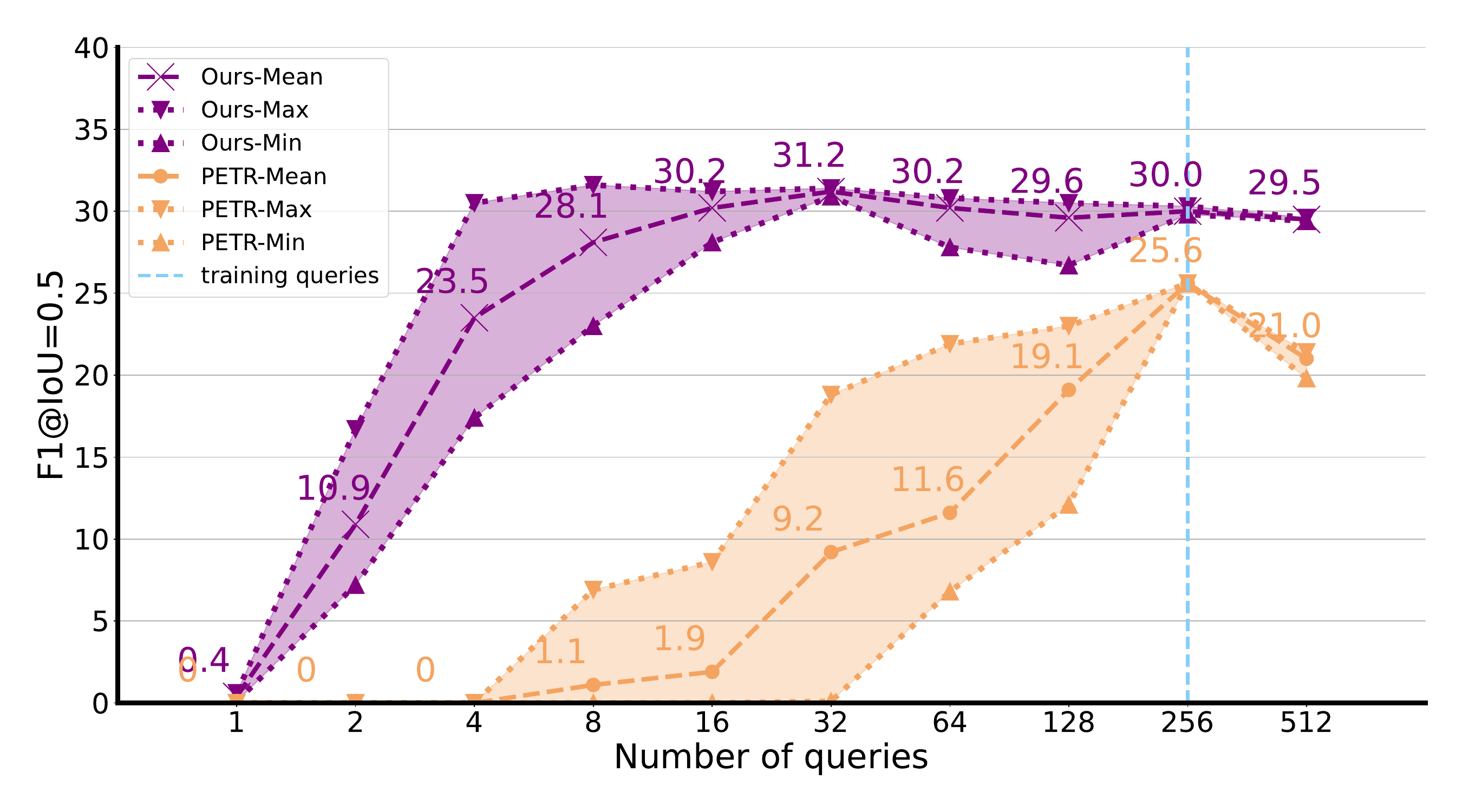}
       \caption{
            \textbf{Varying the number of queries at inference.}
            We run each setting 6 times and report the mean/max/min performance.
           }
       \label{fig:query}
    \vspace{-0.4cm}
\end{figure}

\mypar{Dynamic queries.}
The benefit of transformer architectures is that they can generalize to varying set lengths and are not tied to fixed-resolution inputs. 
For 3D tasks, this is important as one might want to query models with varying numbers of 3D points depending on the resolution of the scene.
But how robust are methods to varying numbers of queries?
We compare \shortname to PETR~\cite{liuPETRPositionEmbedding2022} to demonstrate their efficacy when queried dynamically.
We train both PETR and our model with 256 queries and test the model with different numbers of queries.
Because performance is affected by point distribution, especially in the case of fewer queries, we run each setting 6 times and report the mean/max/min performance in Fig.~\ref{fig:query}. 
When testing with 256 queries (same as during training), we test once with the same distribution of points as during training.
We note that our model is significantly more robust to dynamic queries. 
When the number of queries deviates from training, PETR's performance drops.
This suggests that PETR is prone to overfitting the training distribution.
This connects to our findings from Table~\ref{tab:scannet_view3_ablation} and Fig.~\ref{fig:attention}.
By encoding only geometric cues in the queries, the model focuses on local regions and cannot detect far-away objects which is the case when the queries are few.
This is qualitatively shown in Fig.~\ref{fig:attention}(d).

\begin{figure}[t]
    \vspace{-0.2cm}
    \centering
       \includegraphics[width=1.0\linewidth]{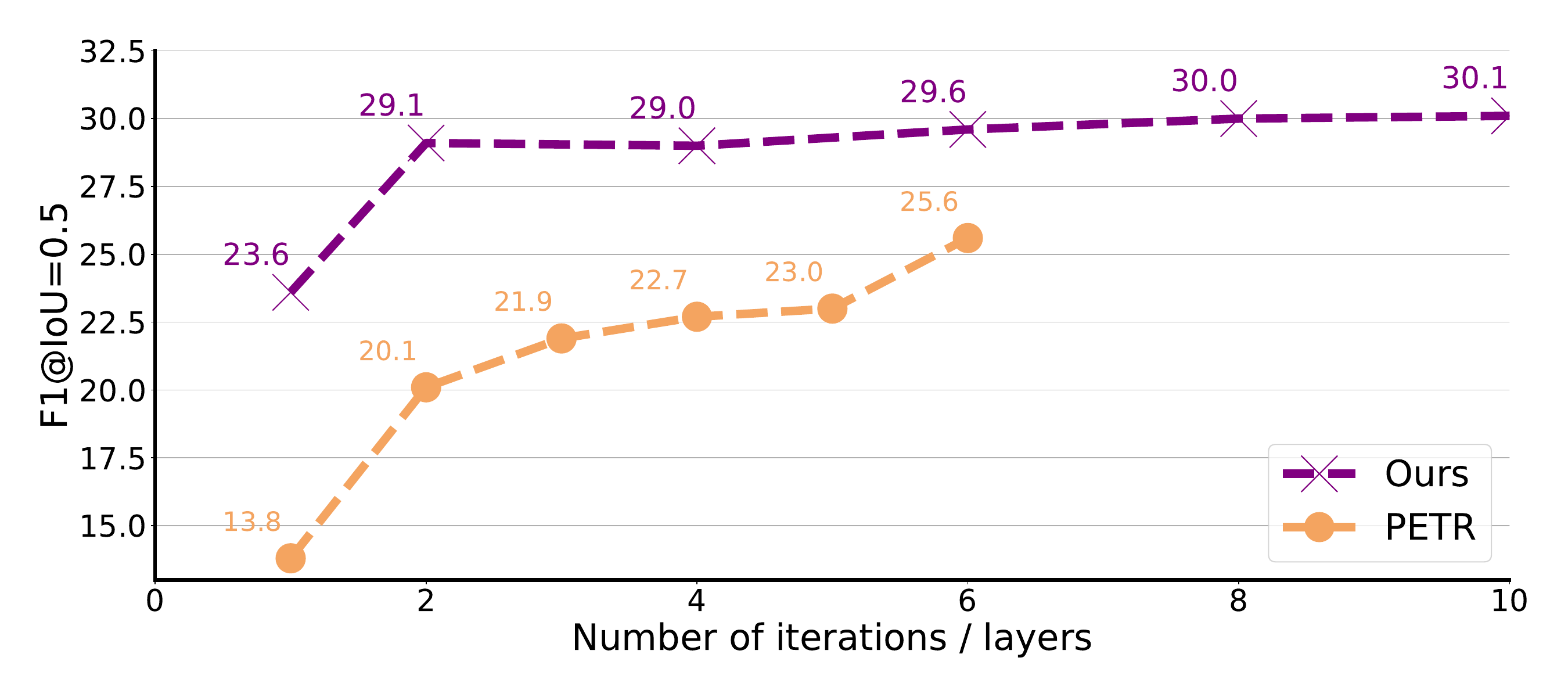}
       \caption{
            \textbf{Varying the number of query updates at inference.} 
            Note PETR doesn't share weights in each layer which means we cannot show its performance for more layers than trained for (6).
           }
       \label{fig:iteration}
       \vspace{-0.4cm}
\end{figure}

\mypar{Recurrent refinement.}
\shortname adopts a recurrent update strategy during training performed by one \shortname layer. 
This design allows us to perform an arbitrary number of refining iterations during inference.
We explore the effect of varying the number of \shortname iterations in Fig.~\ref{fig:iteration} and compare to PETR~\cite{liuPETRPositionEmbedding2022}.
The official implementation of PETR uses 6 layers and does not share weights across layers. 
We report performance from layer 1 to 6.
All models are trained with 256 queries and \shortname is trained to perform 8 iterations.
While \shortname's performance keeps increasing with more layers, it reaches close to maximal performance in just two layers. 
PETR also benefits from updates but lags in performance. 

\mypar{Number of views.}
Our experiments use 3 view video snippets during training and testing.
However, our model can input any number of views during inference.
We show performance for \{1, 2, 3, 5, 7, 9\} views during testing in Table~\ref{tab:views}.
We also compare to PETR~\cite{liuPETRPositionEmbedding2022} and ImVoxelNet~\cite{rukhovichImVoxelNetImageVoxels2021} under the same setting.
All models are trained with 3 views and tested with the number of views indicated.
We observe that PETR and ImVoxelNet perform equally or worse when the number of test views is increased.
\shortname sees a $+1.3\%$ performance boost in F1 by increasing the number of views.

\mypar{Efficiency.}
We report running stats in Table~\ref{tab:efficiency}.
All experiments are conducted on an NVIDIA A5000.
Our model is the most lightweight compared to other baselines.
Our model has an adaptable computation budget at inference which can be controlled with the number of iterations and queries. Based on Fig.~\ref{fig:query} and Fig.~\ref{fig:iteration} we select 32 queries and 2 layers. At these settings, \shortname runs faster (105ms) and detects objects better (F1 $28.2$) than related work.

\section{Limitations}
We observe that our approach can fail when detecting large objects, objects with a similar color 
to the background (\eg, black object in the dark), and objects with heavy occlusion. 
We provide failure cases in the supplementary.

\begin{table}[t]
    \centering
    \resizebox{1.0\columnwidth}{!}{
        \begin{tabular}{cccccccc}
            \Xhline{3\arrayrulewidth}
            Method                                              & Time                    & Test Mem. & Train Mem.      & Param. &  F1$\uparrow$ \\
             & ($ms$)$\downarrow$                    & (GB)$\downarrow$ & (GB)$\downarrow$      & (M)$\downarrow$ & @IoU $>$ 0.5 \\
            \hline
            ImVoxelNet~\cite{rukhovichImVoxelNetImageVoxels2021}  &  127                 & 3.08          & 6.84      &  112.0    & 26.8  \\
            DETR3D~\cite{wangDETR3D3DObject}  &  220                               & \textbf{2.63}      & \textbf{3.93}          &  57.1     & 11.4 \\
            PETR~\cite{liuPETRPositionEmbedding2022}  &  248                               & 3.74       & 8.39          &  131.5   & 25.6 \\
            \multirow{2}{*}{\shortname (ours)}                    & $\textbf{105}^{*}$--                       & \multirow{2}{*}{2.82}   & \multirow{2}{*}{5.83}   & \multirow{2}{*}{\textbf{44.7}}    &   \textbf{28.2 --} \\
                                                 & 290 (below) & & & & \textbf{31.2}
            \\
            \Xhline{3\arrayrulewidth}
            \shortname (ours) & \multicolumn{5}{c}{Time ($ms$) of ours with different settings} \\
            \hline
            \textbf{Num. Iter.} & \textit{1} & \textit{2} & \textit{4} & \textit{6} & \textit{8} \\
            Ours \textit{with 256 quer.} & 180 & 194 & 213 & 252 & 290 \\
            \hline
            \textbf{Num. Quer.} & \textit{16} & \textit{32} & \textit{64} & \textit{128} & \textit{256} \\
            Ours \textit{with 8 iter.} & 156 & 167 & 171 & 227 & 290 \\
            \Xhline{3\arrayrulewidth}
        \end{tabular} }
    \caption{
    \textbf{Efficiency.} The reported inference time, memory consumption, parameter counts and F1 scores show that \shortname is fastest and achieves the highest performance.
    ${^{*}}$\shortname achieves an inference time of 105ms with 32 queries and 2 layers.
    }
    \label{tab:efficiency}
\end{table}
\section{Conclusion}
We introduce \shortname for multi-view 3D object detection.
\shortname's key idea is to leverage the pixel-aligned queries initialized from reference points in 3D space and to update their 3D locations layer-by-layer to the 3D object center with recurrent cross-attention operations.
Our design enables \shortname to encode the 3D-to-2D correspondences and capture global contextual information of the input images, as we demonstrate with our qualitative analysis.
Experiments show that \shortname outperforms prior best methods, learns and detects faster, is more robust to distribution shifts in reference points, can leverage additional input views, while inference compute can be adapted by changing the number of recurrent iterations.
\FloatBarrier

{\small
\bibliographystyle{ieee_fullname}
\bibliography{egbib.bib}

\begin{thebibliography}{10}\itemsep=-1pt

\bibitem{AugmentedRealityApple}
\href{https://developer.apple.com/augmented-reality/}{Augmented {{Reality}}
  with {{ARKit}}- {{Apple Developer}}}.

\bibitem{avetisyanScan2CADLearningCAD2018}
Armen Avetisyan, Manuel Dahnert, Angela Dai, Manolis Savva, Angel~X. Chang, and
  Matthias Niessner.
\newblock Scan2cad: Learning cad model alignment in rgb-d scans.
\newblock In {\em CVPR}, 2019.

\bibitem{arkitscenes}
Gilad Baruch, Zhuoyuan Chen, Afshin Dehghan, Tal Dimry, Yuri Feigin, Peter Fu,
  Thomas Gebauer, Brandon Joffe, Daniel Kurz, Arik Schwartz, and Elad Shulman.
\newblock Arkitscenes - a diverse real-world dataset for 3d indoor scene
  understanding using mobile rgb-d data.
\newblock In {\em NeurIPS}, 2021.

\bibitem{brazilKinematic3DObject2020}
Garrick Brazil, Gerard Pons-Moll, Xiaoming Liu, and Bernt Schiele.
\newblock Kinematic 3d object detection in monocular video.
\newblock In {\em ECCV}, 2020.

\bibitem{camposORBSLAM3AccurateOpenSource2020}
Carlos Campos, Richard Elvira, Juan J.~G{\'o}mez Rodr{\'i}guez, Jos{\'e} M.~M.
  Montiel, and Juan~D. Tard{\'o}s.
\newblock {{ORB}}-{{SLAM3}}: {{An Accurate Open}}-{{Source Library}} for
  {{Visual}}, {{Visual}}-{{Inertial}} and {{Multi}}-{{Map SLAM}}.
\newblock {\em ArXiv}, 2020.

\bibitem{carionEndtoEndObjectDetection2020}
Nicolas Carion, Francisco Massa, Gabriel Synnaeve, Nicolas Usunier, Alexander
  Kirillov, and Sergey Zagoruyko.
\newblock End-to-end object detection with transformers.
\newblock In {\em ECCV}, 2020.

\bibitem{chen2016monocular}
Xiaozhi Chen, Kaustav Kundu, Ziyu Zhang, Huimin Ma, Sanja Fidler, and Raquel
  Urtasun.
\newblock Monocular 3d object detection for autonomous driving.
\newblock In {\em CVPR}, 2016.

\bibitem{daiScanNetRichlyannotated3D2017}
Angela Dai, Angel~X Chang, Manolis Savva, Maciej Halber, Thomas Funkhouser, and
  Matthias Nie{\ss}ner.
\newblock Scannet: Richly-annotated 3d reconstructions of indoor scenes.
\newblock In {\em CVPR}, 2017.

\bibitem{dai2017bundlefusion}
Angela Dai, Matthias Nie{\ss}ner, Michael Zollh{\"o}fer, Shahram Izadi, and
  Christian Theobalt.
\newblock Bundlefusion: Real-time globally consistent 3d reconstruction using
  on-the-fly surface reintegration.
\newblock {\em ToG}, 2017.

\bibitem{dosovitskiy2020image}
Alexey Dosovitskiy, Lucas Beyer, Alexander Kolesnikov, Dirk Weissenborn,
  Xiaohua Zhai, Thomas Unterthiner, Mostafa Dehghani, Matthias Minderer, Georg
  Heigold, Sylvain Gelly, et~al.
\newblock An image is worth 16x16 words: Transformers for image recognition at
  scale.
\newblock In {\em ICLR}, 2020.

\bibitem{cc3dt}
Tobias Fischer, Yung-Hsu Yang, Suryansh Kumar, Min Sun, and Fisher Yu.
\newblock Cc-3dt: Panoramic 3d object tracking via cross-camera fusion.
\newblock In {\em CoRL}, 2022.

\bibitem{goyal2017accurate}
Priya Goyal, Piotr Doll{\'a}r, Ross Girshick, Pieter Noordhuis, Lukasz
  Wesolowski, Aapo Kyrola, Andrew Tulloch, Yangqing Jia, and Kaiming He.
\newblock Accurate, large minibatch sgd: Training imagenet in 1 hour.
\newblock {\em ArXiv}, 2017.

\bibitem{halko2011finding}
Nathan Halko, Per-Gunnar Martinsson, and Joel~A Tropp.
\newblock Finding structure with randomness: Probabilistic algorithms for
  constructing approximate matrix decompositions.
\newblock {\em SIAM review}, 2011.

\bibitem{hartley2003}
Richard Hartley and Andrew Zisserman.
\newblock {\em Multiple view geometry in computer vision}.
\newblock Cambridge university press, 2003.

\bibitem{heMaskRCNN2018}
Kaiming He, Georgia Gkioxari, Piotr Doll{\'a}r, and Ross Girshick.
\newblock Mask r-cnn.
\newblock In {\em ICCV}, 2017.

\bibitem{heDeepResidualLearning2015}
Kaiming He, Xiangyu Zhang, Shaoqing Ren, and Jian Sun.
\newblock Deep residual learning for image recognition.
\newblock In {\em CVPR}, 2016.

\bibitem{nerfrpn}
Benran Hu, Junkai Huang, Yichen Liu, Yu-Wing Tai, and Chi-Keung Tang.
\newblock Nerf-rpn: A general framework for object detection in nerfs.
\newblock In {\em CVPR}, 2023.

\bibitem{Hu2019Mono3DT}
Hou-Ning Hu, Qi-Zhi Cai, Dequan Wang, Ji Lin, Min Sun, Philipp Krähenbühl,
  Trevor Darrell, and Fisher Yu.
\newblock Joint monocular 3d vehicle detection and tracking.
\newblock In {\em ICCV}, 2019.

\bibitem{huangBEVDet4DExploitTemporal2022}
Junjie Huang and Guan Huang.
\newblock {BEVDet4D}: {Exploit} {Temporal} {Cues} in {Multi}-camera {3D}
  {Object} {Detection}.
\newblock {\em ArXiv}, 2022.

\bibitem{huangBEVDetHighperformanceMulticamera2022}
Junjie Huang, Guan Huang, Zheng Zhu, Yun Ye, and Dalong Du.
\newblock {BEVDet}: {High}-performance {Multi}-camera {3D} {Object} {Detection}
  in {Bird}-{Eye}-{View}.
\newblock {\em ArXiv}, 2022.

\bibitem{kuhnHungarianMethodAssignment2010}
Harold~W Kuhn.
\newblock The hungarian method for the assignment problem.
\newblock {\em Naval research logistics quarterly}, 1955.

\bibitem{liODAMObjectDetection}
Kejie Li, Daniel DeTone, Steven Chen, Minh Vo, Ian Reid, Hamid Rezatoﬁghi,
  Chris Sweeney, Julian Straub, and Richard Newcombe.
\newblock {ODAM}: {Object} {Detection}, {Association}, and {Mapping} using
  {Posed} {RGB} {Video}.
\newblock In {\em ICCV}, 2021.

\bibitem{liMOLTRMultipleObject2020}
Kejie Li, Hamid Rezatofighi, and Ian Reid.
\newblock Moltr: Multiple object localization, tracking and reconstruction from
  monocular rgb videos.
\newblock {\em RA-L}, 2021.

\bibitem{liFroDODetections3D2020}
Kejie Li, Martin Rünz, Meng Tang, Lingni Ma, Chen Kong, Tanner Schmidt, Ian
  Reid, Lourdes Agapito, Julian Straub, Steven Lovegrove, and Richard Newcombe.
\newblock {FroDO}: {From} {Detections} to {3D} {Objects}.
\newblock In {\em CVPR}, 2020.

\bibitem{liBEVFormerLearningBird2022}
Zhiqi Li, Wenhai Wang, Hongyang Li, Enze Xie, Chonghao Sima, Tong Lu, Qiao Yu,
  and Jifeng Dai.
\newblock {BEVFormer}: {Learning} {Bird}'s-{Eye}-{View} {Representation} from
  {Multi}-{Camera} {Images} via {Spatiotemporal} {Transformers}.
\newblock In {\em ECCV}, 2022.

\bibitem{linFeaturePyramidNetworks2017}
Tsung-Yi Lin, Piotr Dollár, Ross Girshick, Kaiming He, Bharath Hariharan, and
  Serge Belongie.
\newblock Feature {Pyramid} {Networks} for {Object} {Detection}.
\newblock In {\em CVPR}, 2017.

\bibitem{liuPETRPositionEmbedding2022}
Yingfei Liu, Tiancai Wang, Xiangyu Zhang, and Jian Sun.
\newblock {PETR}: {Position} {Embedding} {Transformation} for {Multi}-{View}
  {3D} {Object} {Detection}.
\newblock In {\em ECCV}, 2022.

\bibitem{liuPETRv2UnifiedFramework2022}
Yingfei Liu, Junjie Yan, Fan Jia, Shuailin Li, Aqi Gao, Tiancai Wang, Xiangyu
  Zhang, and Jian Sun.
\newblock {PETRv2}: {A} {Unified} {Framework} for {3D} {Perception} from
  {Multi}-{Camera} {Images}.
\newblock {\em ICCV}, 2023.

\bibitem{loshchilov2016sgdr}
Ilya Loshchilov and Frank Hutter.
\newblock Sgdr: Stochastic gradient descent with warm restarts.
\newblock In {\em ICLR}, 2017.

\bibitem{loshchilov2017decoupled}
Ilya Loshchilov and Frank Hutter.
\newblock Decoupled weight decay regularization.
\newblock In {\em ICLR}, 2019.

\bibitem{maninisVid2CADCADModel2022}
Kevis-Kokitsi Maninis, Stefan Popov, Matthias Nie{\ss}ner, and Vittorio
  Ferrari.
\newblock Vid2cad: Cad model alignment using multi-view constraints from
  videos.
\newblock {\em TPAMI}, 2022.

\bibitem{mildenhall2020nerf}
Ben Mildenhall, Pratul~P. Srinivasan, Matthew Tancik, Jonathan~T. Barron, Ravi
  Ramamoorthi, and Ren Ng.
\newblock Nerf: Representing scenes as neural radiance fields for view
  synthesis.
\newblock In {\em ECCV}, 2020.

\bibitem{mildenhallNeRFRepresentingScenes2020}
Ben Mildenhall, Pratul~P. Srinivasan, Matthew Tancik, Jonathan~T. Barron, Ravi
  Ramamoorthi, and Ren Ng.
\newblock Nerf: Representing scenes as neural radiance fields for view
  synthesis.
\newblock In {\em ECCV}, 2020.

\bibitem{Paszke_PyTorch_An_Imperative_2019}
Adam Paszke, Sam Gross, Francisco Massa, Adam Lerer, James Bradbury, Gregory
  Chanan, Trevor Killeen, Zeming Lin, Natalia Gimelshein, Luca Antiga, Alban
  Desmaison, Andreas Kopf, Edward Yang, Zachary DeVito, Martin Raison, Alykhan
  Tejani, Sasank Chilamkurthy, Benoit Steiner, Lu Fang, Junjie Bai, and Soumith
  Chintala.
\newblock {PyTorch: An Imperative Style, High-Performance Deep Learning
  Library}.
\newblock In {\em NeurIPS}, 2019.

\bibitem{qinGeneralOptimizationbasedFramework2019a}
Tong Qin, Jie Pan, Shaozu Cao, and Shaojie Shen.
\newblock A {{General Optimization}}-based {{Framework}} for {{Local Odometry
  Estimation}} with {{Multiple Sensors}}.
\newblock {\em ArXiv}, 2019.

\bibitem{rukhovichImVoxelNetImageVoxels2021}
Danila Rukhovich, Anna Vorontsova, and Anton Konushin.
\newblock Imvoxelnet: Image to voxels projection for monocular and multi-view
  general-purpose 3d object detection.
\newblock In {\em WACV}, 2022.

\bibitem{saitoPIFuPixelAlignedImplicit2019}
Shunsuke Saito, Zeng Huang, Ryota Natsume, Shigeo Morishima, Angjoo Kanazawa,
  and Hao Li.
\newblock Pifu: Pixel-aligned implicit function for high-resolution clothed
  human digitization.
\newblock In {\em ICCV}, 2019.

\bibitem{sunNeuralReconRealTimeCoherent2021}
Jiaming Sun, Yiming Xie, Linghao Chen, Xiaowei Zhou, and Hujun Bao.
\newblock Neuralrecon: Real-time coherent 3d reconstruction from monocular
  video.
\newblock In {\em CVPR}, 2021.

\bibitem{szeliski2022computer}
Richard Szeliski.
\newblock {\em Computer vision: algorithms and applications}.
\newblock Springer Nature, 2022.

\bibitem{Tulsiani_2017_CVPR}
Shubham Tulsiani, Tinghui Zhou, Alexei~A. Efros, and Jitendra Malik.
\newblock Multi-view supervision for single-view reconstruction via
  differentiable ray consistency.
\newblock In {\em CVPR}, 2017.

\bibitem{tyszkiewiczRayTran3DPose2022}
Michał~J. Tyszkiewicz, Kevis-Kokitsi Maninis, Stefan Popov, and Vittorio
  Ferrari.
\newblock {RayTran}: {3D} pose estimation and shape reconstruction of multiple
  objects from videos with ray-traced transformers.
\newblock In {\em ECCV}, 2022.

\bibitem{vaswani2017attention}
Ashish Vaswani, Noam Shazeer, Niki Parmar, Jakob Uszkoreit, Llion Jones,
  Aidan~N Gomez, {\L}ukasz Kaiser, and Illia Polosukhin.
\newblock Attention is all you need.
\newblock In {\em NeurIPS}, 2017.

\bibitem{wangMonocular3DObject2022}
Tai Wang, Jiangmiao Pang, and Dahua Lin.
\newblock Monocular {3D} {Object} {Detection} with {Depth} from {Motion}.
\newblock In {\em ECCV}, 2022.

\bibitem{wangDETR3D3DObject}
Yue Wang, Vitor Guizilini, Tianyuan Zhang, Yilun Wang, Hang Zhao, , and
  Justin~M. Solomon.
\newblock Detr3d: 3d object detection from multi-view images via 3d-to-2d
  queries.
\newblock In {\em CoRL}, 2021.

\bibitem{wheatstone}
Charles Wheatstone.
\newblock Contributions to the physiology of vision.—part the first. on some
  remarkable and hitherto unobserved phenomena of binocular vision.
\newblock In {\em Abstracts of the Papers Printed in the Philosophical
  Transactions of the Royal Society of London}, 1843.

\bibitem{xiePlanarReconRealtime3D2022}
Yiming Xie, Matheus Gadelha, Fengting Yang, Xiaowei Zhou, and Huaizu Jiang.
\newblock {PlanarRecon}: Real-time {3D} plane detection and reconstruction from
  posed monocular videos.
\newblock In {\em CVPR}, 2022.

\bibitem{xu2023nerfdet}
Chenfeng Xu, Bichen Wu, Ji Hou, Sam Tsai, Ruilong Li, Jialiang Wang, Wei Zhan,
  Zijian He, Peter Vajda, Kurt Keutzer, and Masayoshi Tomizuka.
\newblock Nerf-det: Learning geometry-aware volumetric representation for
  multi-view 3d object detection.
\newblock In {\em ICCV}, 2023.

\bibitem{yifan2022input}
Wang Yifan, Carl Doersch, Relja Arandjelovi{\'c}, Joao Carreira, and Andrew
  Zisserman.
\newblock Input-level inductive biases for 3d reconstruction.
\newblock In {\em CVPR}, 2022.

\bibitem{yinCenterbased3DObject2020}
Tianwei Yin, Xingyi Zhou, and Philipp Krahenbuhl.
\newblock Center-based 3d object detection and tracking.
\newblock In {\em CVPR}, 2021.

\bibitem{zhouContinuityRotationRepresentations2018}
Yi Zhou, Connelly Barnes, Lu Jingwan, Yang Jimei, and Li Hao.
\newblock On the continuity of rotation representations in neural networks.
\newblock In {\em CVPR}, 2019.

\end{thebibliography}
}

\end{document}